%% file: main.tex
\documentclass[sigconf]{aamas} 

\renewcommand\footnotetextcopyrightpermission[1]{} 

\usepackage{balance}
\usepackage{lineno}
\usepackage{latexsym}
\usepackage{amsmath}
\usepackage{amsthm}
\usepackage{booktabs}
\usepackage{multirow}
\usepackage{enumitem}
\usepackage{graphicx}
\usepackage{color}
\usepackage{tcolorbox}
\usepackage{bbold}
\usepackage{hyperref}
\usepackage{float}
\usepackage{xcolor}
\usepackage[table]{xcolor}
\usepackage{array}
\usepackage{anyfontsize}
\definecolor{lightgray}{HTML}{F5F5F5}
\definecolor{tableshade}{rgb}{0.94, 1.0, 0.94}
\usepackage{fontawesome}

\title{\texorpdfstring{BAPPA: \textbf{\underline{B}}enchmarking \textbf{\underline{A}}gents, 
\textbf{\underline{P}}lans, and \textbf{\underline{P}}ipelines for 
\textbf{\underline{A}}utomated Text-to-SQL Generation}%
{BAPPA: Bridging Agents, Plans, and Pipelines for Automated Text-to-SQL Generation}}

\author{
\textbf{Fahim Ahmed\textsuperscript{1}\footnotemark[1]},\,
\textbf{Md Mubtasim Ahasan\textsuperscript{1}},\,
\textbf{Jahir Sadik Monon\textsuperscript{1,2}},\,
\textbf{Muntasir Wahed\textsuperscript{3}},\\
\textbf{M Ashraful Amin\textsuperscript{1}},\,
\textbf{A K M Mahbubur Rahman\textsuperscript{1}},\,
\textbf{Amin Ahsan Ali\textsuperscript{1}}\\[1ex]
\textsuperscript{1} Center for Computational \& Data Sciences, Independent University, Bangladesh\\
\textsuperscript{2} University of Massachusetts Amherst \quad
\textsuperscript{3} University of Illinois Urbana-Champaign
}

\begin{abstract}
Text-to-SQL systems provide a natural language interface that can enable even laymen to access information stored in databases. However, existing Large Language Models (LLM) struggle with SQL generation from natural instructions due to large schema sizes and complex reasoning. Prior work often focuses on complex, somewhat impractical pipelines using flagship models, while smaller, efficient models remain overlooked. In this work, we explore three multi-agent LLM pipelines, with systematic performance benchmarking across a range of small to large open-source models: (1) Multi-agent discussion pipeline, where agents iteratively critique and refine SQL queries, and a judge synthesizes the final answer; (2) Planner-Coder pipeline, where a thinking model planner generates stepwise SQL generation plans and a coder synthesizes queries; and (3) Coder-Aggregator pipeline, where multiple coders independently generate SQL queries, and a reasoning agent selects the best query. 
Experiments on the Bird-Bench Mini-Dev set reveal that Multi-Agent discussion can improve small model performance, with up to 10.6\% increase in Execution Accuracy for Qwen2.5-7b-Instruct seen after three rounds of discussion. Among the pipelines, the LLM Reasoner-Coder pipeline yields the best results, with DeepSeek-R1-32B and QwQ-32B planners boosting Gemma 3 27B IT accuracy from 52.4\% to the highest score of 56.4\%.
Codes are available at \href{https://github.com/treeDweller98/bappa-sql}{\texttt{\textcolor[RGB]{0, 102, 204}{github.com/treeDweller98/bappa-sql}}}.
\end{abstract}

\newcommand{\BibTeX}{\rm B\kern-.05em{\sc i\kern-.025em b}\kern-.08em\TeX}

\newcolumntype{g}{>{\columncolor{tableshade}}c} 
\newcommand{\newtext}[1]{\textcolor{black}{#1}} 

\begin{document}


\pagestyle{fancy}
\fancyhead{}


\maketitle 

\renewcommand{\thefootnote}{\fnsymbol{footnote}}
\footnotetext{*Corresponding author: \texttt{aca.ahmed.fahim@gmail.com}}

\section{Introduction}
\input{sections/intro}

\section{Related Works}
\input{sections/literature}

\section{Methodology}
\input{sections/methodology}

\section{Experimental Setup}
\input{sections/setup}

\section{Experimental Results and Discussion}
\input{sections/results}

\section{Conclusion}

We explored how Multi-Agent Discussion and reasoning-based planning can enhance Text-to-SQL generation using small and mid-sized open-source language models. Our evaluations on BIRD-SQL and Spider show that multi-agent discussion benefits smaller and mid-scale models the most, with up to +9.6 EX improvement for Qwen2.5-Coder-14B-Instruct on BIRD. Reasoning-based planning further boosts performance, especially for Gemma 3 27B IT, which reached 56.4 EX on BIRD and 71.7 EX on Spider when guided by strong joint planners. While large coders show diminishing returns or minor gains, smaller and mid-tier models benefit substantially from structured reasoning and aggregation. The Coder–Aggregator pipeline further improves reliability, reaching 54.4 EX on BIRD and 75.1 EX on Spider. These results underscore the potential of collaborative, multi-agent pipelines to enhance Text-to-SQL.

\clearpage
\bibliographystyle{ACM-Reference-Format} 
\bibliography{main}

\appendix
\input{sections/appendix}


\end{document}

%% file: sections/intro.tex
In recent years, Large Language Models (LLMs) have demonstrated strong capabilities in translating natural language into code \citep{rp1, rp2}. Text-to-SQL tasks, which focus on converting natural language into executable SQL queries \citep{birdbench}, similarly offer strong potential to automate query construction and reduce manual effort. However, limited attention has been paid to efficient approaches using smaller or open models. Most existing work focuses on scaling up model size, which demands more resources and limits accessibility \citep{small-llm-sql}, or on developing proprietary models and complex systems that introduce challenges in cost, privacy, and adaptability for real-world deployment \citep{rt16}.

\textbf{Challenge 1: The potential of multi-agent LLM pipelines for direct Text-to-SQL generation remains underexplored.} Existing Text-to-SQL systems largely depend on complex, fine-tuned pipelines with specialized models for subtasks such as schema linking or context encoding \citep{sql-llm-survey, structure-guided-sql, finetune-sql-context}. While multi-agent pipelines, where LLMs collaborate through role specialization and critique, have shown strong gains in reasoning and generation quality across domains \citep{rp1, rp2, rp5}, their application to SQL generation remains relatively limited. In particular, “LLMs-as-Judge” frameworks, where agents evaluate and refine each other’s outputs \citep{rp7, rp8}, and role-based designs like “Planner” and “Coder” architectures \citep{plan1, plan2} demonstrate the promise of structured collaboration. However, the potential of collaborative, society-of-mind multi-agent approaches remains underexplored, and such frameworks have not yet been applied to the Text-to-SQL task, despite their promise for enhancing reasoning and generation quality.

\textbf{Challenge 2: Systematic benchmarking of recent open-source LLMs across model scales remains missing.} Prior work has mostly relied on proprietary models such as GPT, Claude, and Gemini via API calls, raising privacy risks from exposing sensitive database information \citep{privacy_sql} and high cost concerns \citep{api-cost-sql}. Yet for real-world deployment, open-source and smaller models are crucial to ensure privacy and adaptability in resource-constrained environments. Meanwhile, families such as Qwen-2, Gemma, CodeLLaMA, Granite, DeepSeek, and StarCoder are advancing rapidly and show strong capabilities across tasks. Despite this progress, their relative strengths for Text-to-SQL remain unclear, leaving a critical gap.  

\textbf{We address the first challenge by proposing a suite of multi-agent LLM pipelines for Text-to-SQL.} Our contribution is the design of three strategies that decompose SQL generation into agentic subtasks, planning, critique, and aggregation, enabling specialized agents to collaborate and refine query quality. The \textit{\textbf{(i) Multi-Agent Discussion pipeline}} engages three agents with distinct personas (Simple, Technical, Thinker) that iteratively exchange critiques and revisions across rounds, guided by a Judge agent who synthesizes the final SQL through consensus. The \textit{\textbf{(ii) Planner–Coder pipeline}} separates reasoning and execution: a Planner analyzes the schema and user query to produce a structured plan, which a Coder then implements as SQL, allowing transparent reasoning and plan reuse. The \textit{\textbf{(iii) Coder–Aggregator pipeline}} has multiple Coders independently propose candidate SQL queries with reasoning traces, while an Aggregator evaluates and integrates them into a unified output. This design unifies diverse query perspectives and reduces the impact of individual agent errors.

\textbf{We address the second challenge by conducting a systematic evaluation of recent open-source LLMs across diverse families and scales.} Our benchmark includes instruction-tuned general-purpose models such as Qwen2 (7B, 14B, 32B), Gemma 3 (4B, 12B, 27B), and Granite 3 (8B), as well as code-specialized variants including Qwen2-Coder, CodeLLaMA (7B, 13B, 34B), DeepSeek-Coder, and StarCoder-15B. We further examine emerging models such as DeepSeek-R1-Qwen (7B–32B), QwQ-32B, and DeepSeek-V2-Lite-Chat. This evaluation enables comparison across general and code-optimized reasoning, revealing trends in performance and efficiency. Our study establishes a foundation for open, cost-efficient Text-to-SQL systems and demonstrates that multi-agent collaboration can deliver strong results even with smaller models.


\textbf{Contribution}. We summarize our contributions: 
\begin{itemize}
    \item We conduct an extensive evaluation of Text-to-SQL capabilities across 24 open-source LLMs (4B–34B), spanning diverse families and specializations, and establish a strong foundation for open, cost-efficient Text-To-SQL systems.
    
    \item We present the first systematic exploration of multi-agent LLM pipelines for Text-to-SQL generation, introducing and benchmarking three novel designs: \textit{Multi-Agent Discussion}, \textit{Planner–Coder}, and \textit{Coder–Aggregator}

    \item We show that reasoning-focused models can substantially improve SQL generation quality by serving as planners or aggregators, enabling smaller LLMs to achieve performance on par with significantly larger models.
\end{itemize}

%% file: sections/literature.tex
Few works explore complex schema linking for reducing irrelevant input noise and enhancing query generation. RSL-SQL \citep{rp11} introduces a robust schema linking framework to improve performance on the BIRD benchmark, while MCS-SQL \citep{rt12} enhances SQL generation on Spider through multi-prompt strategies and fine-grained schema modeling. However, \citet{rp13} show that modern LLMs can often identify relevant schema components implicitly when full schemas fit within the context window, suggesting explicit linking may be redundant in some setups. Other efforts focus on model adaptation, such as CodeS \citep{code-sql}, pre-trained on SQL data for domain specialization, and DTS-SQL \citep{rt16} and Reasoning-SQL \citep{grpo-sql}, which fine-tune smaller open-source models. The latter demonstrates that reasoning-aware supervision can make compact models competitive with larger closed-source ones.

DTS-SQL \citep{rt16} introduces a two-stage fine-tuning approach that decomposes the Text-to-SQL task into simpler sub-tasks, enabling 7B-scale open-source models to approach the performance of larger commercial ones. DIN-SQL \citep{rt15} similarly integrates intermediate reasoning steps, while TA-SQL \citep{rt14} aligns tasks using semantically similar examples to mitigate hallucinations. MCTS-SQL \citep{mcts-sql} applies Monte Carlo Tree Search as a self-refinement mechanism to iteratively improve SQL accuracy and consistency on challenging benchmarks like BIRD and Spider. LLM-based agents have gained traction for modular reasoning and tool use. OpenAgents \citep{openagents-cite} introduces specialized single-agent modules for structured task execution. MAC-SQL \citep{mac-sql} adopts a fixed-role multi-agent setup using proprietary models such as GPT-4, requiring smaller models to be fine-tuned to imitate responses.

\newtext{In contrast, our work develops flexible multi-agent pipelines where roles are dynamically assigned to open-source models without additional fine-tuning. We focus on agent interaction and modular role specialization such as Planner–Coder, Coder–Aggregator, and Multi-Agent Discussion to enable structured SQL generation with both general-purpose and code-focused LLMs.}

%% file: sections/methodology.tex
\begin{figure*}[ht]
\begin{center}
    \includegraphics[width=0.90\linewidth, height=0.5\linewidth]{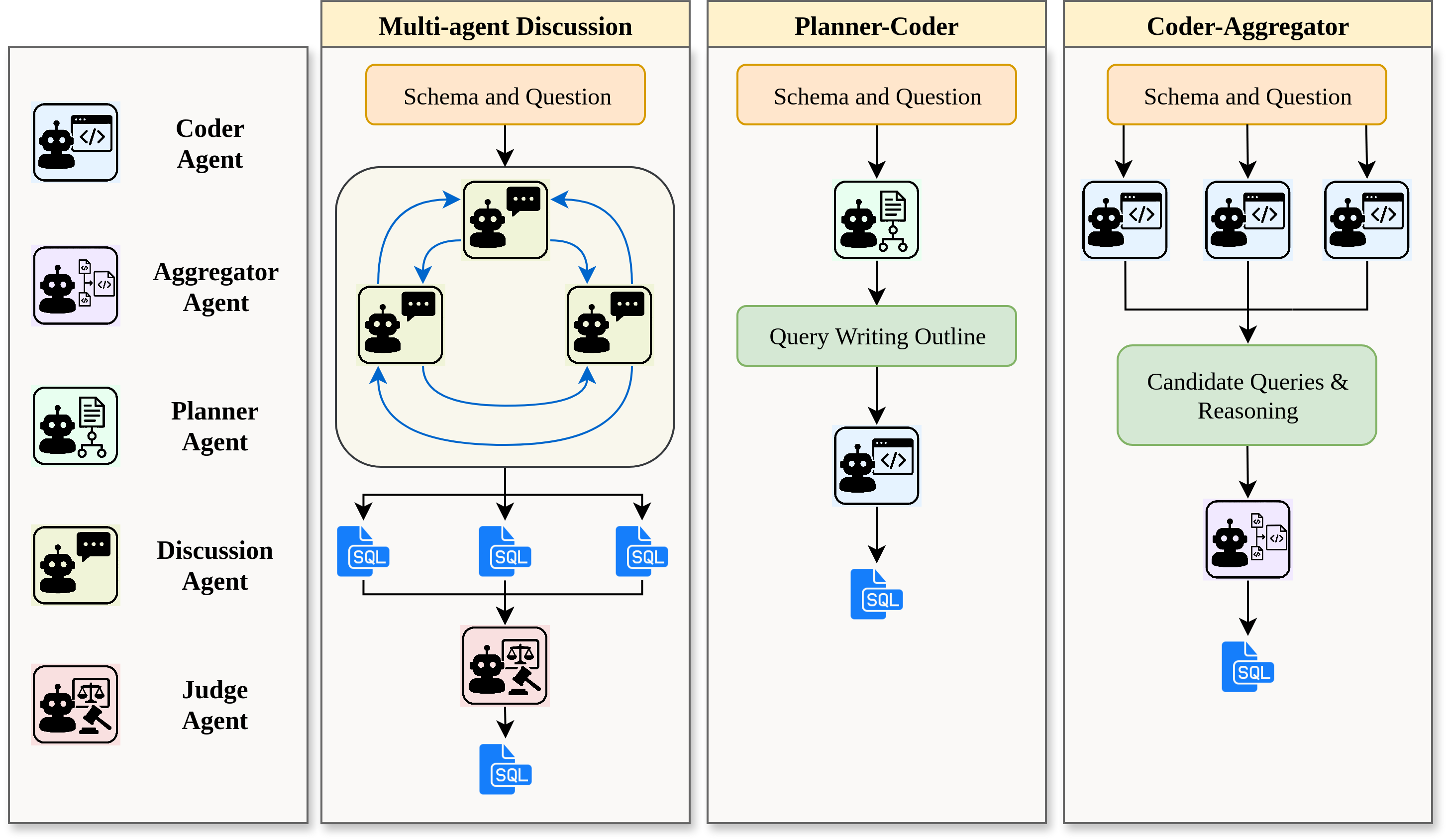}
\end{center}
\Description{Diagram showing three multi-agent pipelines for Text-to-SQL}
\caption{Overview of Proposed Multi-Agent Pipelines for Text-to-SQL. We propose three pipelines: (i) Multi-agent Discussion, where Discussion Agents iteratively critique and refine each other's responses before a final SQL is selected by a Judge Agent; (ii) Planner-Coder, where a Planner Agent generates a step-by-step outline used by a Coder Agent to synthesize SQL; and (iii) Coder-Aggregator, where multiple Coder Agents generate candidate SQL queries, and an Aggregator Agent selects the final output. All pipelines take a schema and question as input.}
\label{fig:pipelines}
\end{figure*}

We propose three multi-agent LLM pipelines for Text-to-SQL generation, emphasizing agent collaboration and coordination. We focus on LLM-only settings to evaluate intrinsic planning, reasoning, and coding abilities without external supervision. As shown in Figure~\ref{fig:pipelines}, the proposed pipelines include \textbf{Multi-Agent Discussion} (§\ref{sec:discussion}), \textbf{Planner–Coder} (§\ref{sec:planner-coder}), and \textbf{Coder–Aggregator} (§\ref{sec:coder-aggregator}).

\subsection{Multi-agent Discussion}
\label{sec:discussion}

The factuality and reasoning abilities of large language models can be enhanced through collaboration. Their tendency toward fallacious reasoning and hallucination can be mitigated by prompting them to consider peer responses when generating their own. We adopt a \textit{Society of Minds}~\cite{improvefact-mad} approach to Text-to-SQL, where three agents engage in multi-round discussion to produce an SQL query. Diverse starter queries are first generated by LLM instances with distinct personas: one favors concise solutions, another gives detailed technical answers, and the third explores alternatives before deciding. Each \textbf{\textit{Discussion Agent}} critiques others’ responses and revises its own, while a \textbf{\textit{Judge Agent}} integrates their outputs into a final SQL query at the end of each round.

Formally, given an input question \( q \) and database schema \( s \), each agent \( a_i \in \{1, 2, 3\} \) generates an initial SQL query \( x_i^{(0)} \) conditioned on its persona \( p_i \):
\[x_i^{(0)} = \text{StarterAgent}_i(s, q \mid p_i).\]

At discussion round \( t = 1, 2, 3 \), each agent observes the outputs of other agents, \( \{x_j^{(t-1)} \mid j \ne i\} \), and generates a revised response:
\[x_i^{(t)} = \text{DiscussionAgent}_i(s, q, \{x_j^{(t-1)}\}_{j \ne i}).\]

A Judge Agent then aggregates the three agent responses into a single verdict query:
\[\hat{x}^{(t)} = \text{JudgeAgent}(s, q, \{x_1^{(t)}, x_2^{(t)}, x_3^{(t)}\}).\]

This process repeats for \( T = 3 \) rounds. The final output \( \hat{x}^{(3)} \) serves as the system’s prediction. By simulating collaborative critique and revision, this framework promotes robustness and mitigates typical failure modes in single-shot generation.

\subsection{Planner-Coder}
\label{sec:planner-coder}

Recent thinking models demonstrate markedly improved reasoning and planning capabilities by leveraging extended test-time computation. We introduce \textbf{\textit{Planner Agents}}—thinking models that generate structured, step-by-step outlines for translating natural language questions into executable SQL. This design explicitly harnesses reflective reasoning to guide subsequent code generation.

Formally, given a natural language question \( q \) and a database schema \( s \), a Planner Agent produces a plan or a set of plans \( z = \{z_1, z_2, \ldots, z_k\} \) that outlines how to answer the question:
\[z = \text{PlannerAgent}(s, q).\]

A Coder Agent then generates the final SQL query \( x \) conditioned on both the schema and the plan(s):
\[x = \text{CoderAgent}(s, q, z).\]

This framework allows us to isolate the contribution of structured intermediate reasoning in SQL generation and study whether diverse planning leads to more robust execution.

\subsection{Coder-Aggregator}
\label{sec:coder-aggregator}

Recent LLMs can generate high-quality SQL queries in a zero-shot setting using chain-of-thought (CoT) reasoning. To further enhance reliability and execution consistency, we employ a two-stage framework: multiple \textbf{\textit{Coder Agents}} independently produce SQL candidates with reasoning traces, while a single \textbf{\textit{Aggregator Agent}} synthesizes the final query by integrating these diverse outputs.

Formally, given a natural language question \( q \) and database schema \( s \), we first generate a set of candidate outputs \( \{(x_i, r_i)\}_{i=1}^n \) from multiple Coder Agents, where \( x_i \) is a candidate SQL query and \( r_i \) its accompanying reasoning:
\[\{(x_i, r_i)\}_{i=1}^n = \text{CoderAgents}(s, q)\]

The Aggregator Agent then generates a final SQL query \( x^* \), conditioned on the schema, question, and the set of candidate outputs:
\[x^* = \text{AggregatorAgent}(s, q, \{(x_i, r_i)\}_{i=1}^n).\]

This setup enables the model to benefit from multiple diverse perspectives during inference, improving factual consistency and execution accuracy through a form of self-critique and consensus.

%% file: sections/setup.tex
This section outlines the experimental framework for evaluating our methods, including the dataset (§\ref{sec:dataset}), evaluation metric (§\ref{sec:metric}), baseline prompt (§\ref{sec:baseline}), and inference setup (§\ref{sec:inference}).

\subsection{Dataset}

\label{sec:dataset}

We conduct experiments on the Spider \cite{spider} Dev and BIRD \cite{birdbench} Mini-Dev SQLite subsets. BIRD Mini-Dev includes 500 natural language–SQL pairs from 11 databases, annotated by difficulty: \textit{simple} (148), \textit{moderate} (250), and \textit{hard} (102). Spider Dev contains 1,034 pairs across 166 databases. Their challenging nature make them suitable for large-scale evaluation in an efficient, cost-effective manner.

\subsection{Metric}
\label{sec:metric}

To evaluate the correctness and quality of the SQL queries generated, we use the following metrics described by \cite{birdbench}:

\textbf{Execution Accuracy (EX)} is the proportion of evaluation examples where the executed results of the predicted and ground-truth SQL queries match. Formally, let \( V_n \) be the result set from the ground-truth query \( Y_n \), and \( \hat{V}_n \) the result set from the predicted query \( \hat{Y}_n \). Then, EX is defined as:

\begin{equation}
    \text{EX} = \frac{\sum_{n=1}^{N} \mathbb{1}(V_n = \hat{V}_n)}{N},
\end{equation}
where \( \mathbb{1}(\cdot) \) is an indicator function:
\begin{equation}
    \mathbb{1}(V = \hat{V}) = 
    \begin{cases}
    1, & \text{if } V = \hat{V} \\
    0, & \text{otherwise}
    \end{cases}
\end{equation}

\textbf{Soft F1-Score.} Unlike EX, this metric offers a more tolerant evaluation by accounting for partial matches and reducing sensitivity to column order or missing values. Given ground-truth and predicted SQL result tables, we compare corresponding rows to count matched ($TP$), prediction-only ($FP$), and gold-only ($FN$) elements, from which precision, recall, and F1 are derived:

\begin{align}
\text{Precision} &= \frac{TP}{TP + FP}, \quad 
\text{Recall} = \frac{TP}{TP + FN}, \\
\text{Soft-F1} &= \frac{2 \times \text{Precision} \times \text{Recall}}{\text{Precision} + \text{Recall}}.
\end{align}

This metric provides a balanced and robust measure of semantic correctness, capturing model performance even without exact structural alignment between predicted and ground-truth results.

\textbf{Reward-based Validation Efficiency Score (R-VES).} This metric jointly evaluates the accuracy and efficiency of correct SQL queries (\(\hat{Y}\), the predicted query, satisfies \(V_n = \hat{V}_n\)) by rewarding faster execution. Let \(\tau = \frac{T_{\text{GT}}}{T_{\text{Pred}}}\) denote the execution time ratio between ground-truth and predicted SQLs, averaged over 100 runs with outliers removed. Then, R-VES is defined as:

\begin{equation}
\text{R-VES} =
\begin{cases}
1.25, & \text{if } \tau \ge 2, \\
1.0,  & \text{if } 1 \le \tau < 2, \\
0.75, & \text{if } 0.5 \le \tau < 1, \\
0.5,  & \text{if } 0.25 \le \tau < 0.5, \\
0.25, & \text{if } \tau < 0.25, \\
0,    & \text{if } \hat{Y} \text{ is incorrect.}
\end{cases}
\label{eq:r-ves}
\end{equation}

\noindent
This formulation rewards correct, efficient SQL generation while maintaining fairness and robustness.

\subsection{Baseline}

\label{sec:baseline}
To improve SQL accuracy and formatting, we use zero-shot Chain-of-Thought (CoT) prompting by appending ``Let's think step by step'' to the input, following \citet{zscot}. This is applied only for non-thinking models, as best practices advise against such prompts for reasoning-capable LLMs \cite{cot-without-prompting}.

\subsection{Inference Setup}

\label{sec:inference}

All experiments used \texttt{vllm} v0.8.5.post1 offline batched inference on either 2\texttimes \textit{L40S 48GB}, 2\texttimes \textit{A40 48GB}, or 1\texttimes \textit{A100 80GB} GPU. We set \texttt{tensor\_parallelism} to 2 (for dual GPUs), \texttt{max\_model\_len} to 16,384, and enable \texttt{enable\_prefix\_caching} and \texttt{enforce\_eager}. Prompts used model's default chat template. Non-thinking models used greedy decoding with \texttt{repetition\_penalty} 1.05; thinking models used stochastic sampling (\texttt{temperature}=0.6, \texttt{top\_p}=0.95, \texttt{top\_k}=30, \texttt{repetition\_penalty}=1.0). We report best results from multiple runs, with random seeds 42, 11, and 98 for reproducibility.

%% file: sections/results.tex
\input{tables/zs-res}

In this section, we present the experimental results evaluating our pipelines on the BIRD-SQL and Spider benchmarks. We analyze the performance of each method relative to a zero-shot baseline and highlight the contribution of discussion and planning in improving SQL generation.
We report results and analysis in the following sections: zero-shot baseline (§\ref{sec:baseline-results}), multi-agent discussion (§\ref{sec:discussion-results}), planner-coder pipeline (§\ref{sec:planner-coder-results}), and coder-aggregator pipeline (§\ref{sec:aggregator-results}).

\input{tables/mad-res}

\subsection{Baseline Zero-Shot}
\label{sec:baseline-results}

Table \ref{tbl:zs-res} reports the results of SQL queries generated by all our evaluated models on the BIRD Mini-Dev and Spider Dev set. The zero-shot prompt shown in Figure \ref{fig:zeroshot_prompt} prevents the use of any external tools or aids, making it a reliable measure of each model’s raw coding capability. All models, except for the reasoning-based ones, were restricted to a maximum generation length of 1024 tokens per question. Thinking models, relying on lengthy internal monologues, were allowed up to 8192 tokens per query.

\textbf{Zero-shot evaluation exposes strong advances among open models.}  
\textbf{Gemma 3 (27B IT)} delivers the best overall results with 52.4 / 78.9 EX, 56.3 / 83.6 Soft~F1, and 46.3 / 75.7 R-VES on BIRD and Spider, setting a new open-source baseline for zero-shot Text-to-SQL generation.  
Qwen2.5-Coder (14B Instruct) follows closely (49.6 EX / 54.1 Soft~F1 / 47.6  R-VES) on BIRD, confirming that targeted code pretraining can rival or surpass much larger models.  
Both families substantially outperform proprietary GPT-4-Turbo (45.8 EX on BIRD), underscoring rapid progress in open instruction-tuned systems.

\textbf{Mid-scale reasoning and code-oriented models perform robustly.}  
DeepSeek R1 and V2 series reach up to 69.3 EX and 78.0 Soft~F1 on Spider, showing that reasoning distillation improves compositional SQL synthesis. Granite 3 and Mistral Small maintain steady midrange accuracy (60–67 EX; 68–77 Soft~F1), offering balanced performance at moderate cost.  
General-purpose instruction models such as Llama 3 and StarCoder remain behind this tier, while older code-focused models like CodeLlama lag significantly, reflecting limited schema understanding and weaker reasoning ability. Notably, CodeLlama-13B often generated malformed or gibberish SQL tokens, resulting in the lowest EX and R-VES scores in the benchmark.

\textbf{Overall, open instruction-tuned and code-specialized LLMs now rival proprietary systems in zero-shot SQL generation.}  
\textbf{Gemma 3 and Qwen2.5-Coder set a new standard for efficient, accessible, and high-accuracy query generation}, illustrating that carefully aligned reasoning and code supervision matter more than raw model size.

\input{tables/plan-res}

\subsection{Multi-agent Discussion}
\label{sec:discussion-results}

\looseness=-1

Table~\ref{tbl:mad-res} presents the SQL generation performance of the Multi-Agent Discussion (MAD) pipeline. The columns report scores for the SQL generated by the judge at the end of each round, enabling us to track how query quality evolves. To ensure agents have adequate context length, we allocate up to 4096 tokens per turn, a limit rarely approached in practice.

\textbf{The Multi-Agent Discussion pipeline yields stable yet modest gains across dialogue rounds, with collaboration refining consistency more than improving accuracy.}  
Across models, EX, Soft~F1, and R-VES metrics generally plateau after Round~2, indicating that early exchanges capture most of the benefit. Well-aligned instruction models already achieve near-maximum quality in Round~1, leaving limited headroom for iterative refinement.

\textbf{High-performing base models sustain their dominance even in multi-agent settings.}  
Gemma 3 (27B IT) achieves the best Round 1 results with 53.0 EX / 55.2 Soft~F1 / 52.9 R-VES on BIRD and 77.6 EX / 82.7 Soft~F1 / 74.6 R-VES on Spider, while Gemma 3 (12B IT) remains close behind on Round 1 (48.4 EX / 52.8 Soft~F1 / 46.0 R-VES).  
This stability suggests that strong single-agent reasoning outweighs additional debate depth.

\textbf{Mid-scale models benefit from discussion.}  
Qwen2.5-Coder-14B-Instruct improves substantially from \textbf{31.4 → 41.0 EX} on BIRD (and from \textbf{35.8 → 45.9 Soft~F1}), showing that critique helps mid-scale coders reconcile reasoning errors. Granite, Ministral, and DeepSeek-Coder models gain modest but consistent 1–2 EX improvements across rounds, with minor boosts in Soft~F1 and R-VES, indicating better structural consistency rather than semantic leaps.

Conversely, \textbf{CodeLlama models remain largely unaffected or even regress slightly across rounds}, reflecting weak schema grounding and poor adaptability to feedback. The 13B variant in particular shows minimal movement (4–5 EX, < 5 Soft~F1, < 5 R-VES), confirming earlier findings of unstable SQL generation.

\textbf{Overall, multi-agent interaction improves the ability of LLMs to generate correct SQL from user requests, though multiple rounds yield only modest gains while requiring substantially more compute than zero-shot prompting.}

\input{tables/pick-res}

\subsection{Planner-Coder}
\label{sec:planner-coder-results}

\looseness=-1
Table~\ref{tbl:planner-coder-bird-spider} reports the SQL generation performance of the tested models when conditioned on plans produced by reasoning-oriented LLMs, specifically using the Planner-Coder pipeline. Each column shows the execution accuracy (EX) when conditioning on plans from a specific Planner model. The rightmost column presents the EX score when plans from both DeepSeek-R1-Distill-Qwen-32B and QwQ-32B are used jointly. Planner Agents were allowed to generate up to 8192 tokens to accommodate detailed planning, while Coder Agents were limited to 1024 tokens.

\textbf{Planner–Coder conditioning substantially enhances SQL generation, especially for weaker coders.}  
Across both BIRD and Spider, leveraging reasoning-oriented plans consistently improves EX, Soft F1, and R-VES, with the largest gains observed when coders are guided by mid- to large-scale Planner agents. CodeLlama and StarCoder models, which perform poorly in zero-shot settings, benefit the most showing BIRD \textbf{EX gains of up to +23 points}. This highlights the critical role of explicit plan supervision in compensating for limited inherent reasoning ability.

\textbf{Joint planning yields additional, though selective, improvements.}  
Combining plans from R1-32B and QwQ-32B produces the highest execution accuracies, including Gemma-3-4B-IT (48.2 EX / 52.8 Soft-F1 / 46.3 VES) and Qwen-2.5-7B~Instruct (45.8 EX / 53.0 Soft~F1 / 43.7 R-VES) on BIRD. On Spider, joint plans further enhance mid-tier coders such as Qwen-2.5-Coder-14B-Instruct (65.7 EX / 73.6 F1 / 62.9~EM), indicating that complementary planner reasoning can refine query structure and correctness beyond what single planners achieve.

\textbf{High-performing coders exhibit smaller, yet consistent, gains.}  
Gemma~3~12B~IT, Gemma~3~27B~IT, and QwQ~32B already achieve strong execution accuracy, so planner conditioning primarily stabilizes outputs rather than driving large improvements. For instance, Gemma~3~12B~IT attains 51.4 EX / 56.5 Soft~F1 / 49.1 R-VES in BIRD with joint plans, showing modest but consistent gain over its single-planner performance.

\textbf{Overall, the Planner–Coder pipeline acts as a targeted reasoning amplifier.}  
While the strongest gains occur for weaker and mid-tier coders, large-planner or joint-planner conditioning consistently improves SQL accuracy, schema alignment, and execution reliability, underscoring the importance of structured planning in multistep code generation.

\subsection{Coder-Aggregator}
\label{sec:aggregator-results}

\looseness=-1
Table~\ref{tbl:pick-res} presents the SQL generation performance of the Coder-Aggregator pipeline, measured using Execution Accuracy (EX). Each row corresponds to a specific coder model size (small, medium, large), while the leftmost rows represent the aggregator models used to review the candidate queries.  

\textbf{The Coder–Aggregator pipeline demonstrates that ensemble style reasoning can consistently enhance SQL accuracy, particularly for smaller and mid-scale coder sets.}  
Across both datasets, aggregators effectively consolidate diverse SQL candidates, producing steady gains in EX, Soft~F1, and R-VES over single-coder baselines. The improvement is most pronounced on BIRD, where the aggregator mitigates schema-matching and execution errors that individual coders often make.

\textbf{Strong reasoning aggregators achieve the best overall performance.}  
QwQ-32B delivers the highest scores using LARGE Coders on BIRD with 54.4 EX / 60.2 Soft F1 / 53.1 R-VES, outperforming all DeepSeek-R1 variants. On Spider, however, \textbf{DeepSeek-R1-Distill-Qwen-14B leads using LARGE Coders} with 75.1 EX / 82.2 Soft F1 / 72.4 R-VES, confirming its strength in compositional and execution reasoning. The 32B variant trails slightly, indicating diminishing returns from larger aggregation capacity once the reasoning model is well-aligned.

\textbf{Cross-scale results reveal complementary behavior between coders and aggregators.}  
Mid-tier aggregators (14B–32B) yield clear gains when reviewing SQLs from small and mid coders, up to $+$10 EX and $+$8 Soft F1, on BIRD, highlighting that the aggregator’s analytical capability matters more than coder size. Large coders benefit less from aggregation, as their initial SQLs already approach upper-bound quality. 

\textbf{Overall, the Coder–Aggregator pipeline shows that collective evaluation strengthens both accuracy and reliability.}  
While the absolute gains over the best single-agent runs are moderate, the consistent increases across EX, Soft F1, and R-VES confirm that structured aggregation is a practical and cost-efficient strategy for enhancing SQL robustness, particularly when operating with small or mid-scale coders.

%% file: tables/zs-res.tex
\begin{table}[H]
    \caption{
    Zero-shot Text-to-SQL baseline results on the BIRD Mini-Dev and Spider Dev sets. Results are obtained using the zero-shot CoT prompting shown in Figure~\ref{fig:zeroshot_prompt}. \textbf{Bold} marks the best-performance, \underline{underlined} denotes the second-best. \textdagger indicates results reported from the official BIRD Mini-Dev github. 
    Overall, \textbf{Gemma 3 and Qwen2.5-Coder models achieve the highest zero-shot execution accuracy}, showing the competitiveness of small open-source models.}
    \label{tbl:zs-res}
    \resizebox{0.47\textwidth}{!}{
    
        \begin{tabular}{lccc|ccc}
        \toprule
        \textbf{Model} & \multicolumn{3}{c|}{\textbf{BIRD}} & \multicolumn{3}{c}{\textbf{SPIDER}} \\
        \cmidrule(lr){2-4} \cmidrule(lr){5-7}
         & \textbf{EX} & \textbf{Soft F1} & \textbf{R-VES} & \textbf{EX} & \textbf{Soft F1} & \textbf{R-VES} \\
        \midrule
        Llama 3 70B Instruct\textdagger & 40.8 & 44.38 & 39.0 & - & - & - \\
        GPT-4\textdagger & 47.8 & 52.69 & 45.91 & - & - & - \\
        GPT-4 Turbo\textdagger & 45.8 & 50.1 & 44.8 & - & - & - \\
        GPT-3.5 Turbo\textdagger & 38.0 & 41.8 & 41.84 & - & - & - \\
        \midrule
        CodeLlama Instruct 7B  & 12.0 & 13.8 & 10.8 & 22.0 & 25.8 & 20.6 \\
        CodeLlama Instruct 13B  & 5.2 & 6.6 & 5.0 & 9.9 & 11.4 & 9.9 \\
        CodeLlama Instruct 34B  & 22.2 & 25.2 & 21.2 & 36.9 & 43.3 & 35.0 \\
        DeepSeek V2 Lite Chat & 15.6 & 18.4 & 14.5 & 40.6 & 48.7 & 37.9 \\
        DeepSeek Coder V2 Lite Instruct & 35.4 & 41.4 & 33.5 & 66.7 & 75.1 & 63.7 \\
        Gemma 3 4B IT & 33.2 & 35.1 & 31.6 & 66.0 & 70.9 & 63.3 \\
        Gemma 3 12B IT & 48.8 & 50.5 & 46.3 & \underline{75.5} & 80.4 & 72.9 \\
        Gemma 3 27B IT & \textbf{52.4} & \textbf{56.3}& 46.3 & \textbf{78.9} & \textbf{83.6} & \textbf{75.7} \\
        Granite 3.2 8B Instruct & 31.6 & 34.5 & 29.3 & 58.9 & 68.1 & 55.5 \\
        Granite 3.3 8B Instruct & 32.6 & 36.1 & 30.0 & 62.5 & 70.9 & 59.1 \\
        Llama 3.1 8B Instruct & 33.6 & 37.9 & 31.6 & 58.3 & 67.7 & 56.1 \\
        Ministral 8B Instruct & 31.8 & 37.4 & 29.9 & 56.0 & 66.1 & 53.3 \\
        Mistral Small 24B Instruct & 44.8 & 52.7 & 42.3 & 67.3 & 76.7 & 64.5 \\
        Qwen 2.5 7B Instruct & 32.2 & 37.0 & 29.8 & 62.0 & 70.4 & 59.5 \\
        Qwen 2.5 14B Instruct & 41.6 & 48.3 & 39.4 & 65.3 & 74.8 & 62.8 \\
        Qwen 2.5 32B Instruct & 43.8 & 50.9 & 41.3 & 69.9 & 79.3 & 67.3 \\
        Qwen 2.5 Coder 7B Instruct & 41.8 & 46.8 & 39.7 & 68.4 & 76.0 & 65.9 \\
        Qwen 2.5 Coder 14B Instruct & \underline{49.6} & 54.1& \textbf{47.6} & 75.3 & \underline{80.7} & \underline{73.1} \\
        Qwen 2.5 Coder 32B Instruct & 49.4 & \underline{56.2}& \underline{47.2} & 72.3 & 79.6 & 70.0 \\
        StarCoder 15B Instruct & 31.6 & 37.1 & 29.9 & 62.4 & 71.6 & 59.8 \\
        DeepSeek R1 Distill Qwen 7B Instruct & 15.4 & 16.5 & 14.8 & 46.1 & 52.6 & 43.8 \\
        DeepSeek R1 Distill Qwen 14B Instruct & 38.0 & 43.1 & 35.9 & 65.5 & 74.3 & 62.6 \\
        DeepSeek R1 Distill Qwen 32B Instruct & 42.4 & 49.8 & 40.4 & 69.3 & 78.0 & 66.9 \\
        QwQ 32B & 47.6 & 54.2 & 44.85 & 63.0 & 69.4 & 60.4 \\
        \bottomrule
    \end{tabular}
    }
\end{table}

%% file: tables/mad-res.tex
\begin{table*}[ht]
\begin{center}
\caption{
Multi-Agent Discussion (MAD) results on the BIRD Mini-Dev and Spider Dev sets. Each column shows the final SQL generated by the Judge agent after all discussion rounds. \textbf{Bold} marks the best performance, and \underline{underlined} denotes the second-best. Overall, \textbf{Gemma 3 (27B IT) and Qwen2.5-Coder (14B)} achieve the strongest multi-agent performance, confirming that structured collaboration yields modest but consistent gains over zero-shot prompting.}

\label{tbl:mad-res}
\resizebox{1\textwidth}{!}{

\begin{tabular}{lccc|ccc|ccc|ccc|ccc|ccc}
\toprule
 \textbf{Model} & \multicolumn{9}{c|}{\textbf{BIRD}} & \multicolumn{9}{c}{\textbf{SPIDER}} \\ 
\cmidrule(lr){2-10} \cmidrule(lr){11-19} 

 & \multicolumn{3}{c}{\textbf{Round 1}} & \multicolumn{3}{c}{\textbf{Round 2}} & \multicolumn{3}{c|}{\textbf{Round 3}} & \multicolumn{3}{c}{\textbf{Round 1}} & \multicolumn{3}{c}{\textbf{Round 2}} & \multicolumn{3}{c}{\textbf{Round 3}} \\
  & \textbf{EX} & \textbf{Soft F1} & \textbf{R-VES} & \textbf{EX} & \textbf{Soft F1} & \textbf{R-VES} & \textbf{EX} & \textbf{Soft F1} & \textbf{R-VES} & \textbf{EX} & \textbf{Soft F1} & \textbf{R-VES} & \textbf{EX} & \textbf{Soft F1} & \textbf{R-VES} & \textbf{EX} & \textbf{Soft F1} & \textbf{R-VES} \\
\midrule

CodeLlama Instruct 7B & 5.4 & 6.2 & \multicolumn{1}{c|}{4.8} & 4.6 & 5.2 & \multicolumn{1}{c|}{4.0} & 4.0 & 4.8 & \multicolumn{1}{c|}{3.7} & 11.9 & 13.9 & \multicolumn{1}{c|}{11.6} & 11.1 & 13.2 & \multicolumn{1}{c|}{10.6} & 10.4 & 11.8 & 10.0 \\
CodeLlama Instruct 13B & 4.6 & 4.7 & \multicolumn{1}{c|}{4.3} & 3.6 & 4.2 & \multicolumn{1}{c|}{3.4} & 4.0 & 4.8 & \multicolumn{1}{c|}{3.7} & 11.0 & 12.6 & \multicolumn{1}{c|}{10.8} & 10.2 & 12.3 & \multicolumn{1}{c|}{9.9} & 9.8 & 11.5 & 9.6 \\
CodeLlama Instruct 34B & 5.8 & 5.9 & \multicolumn{1}{c|}{5.5} & 6.0 & 6.2 & \multicolumn{1}{c|}{5.7} & 3.4 & 3.7 & \multicolumn{1}{c|}{3.1} & 16.6 & 18.3 & \multicolumn{1}{c|}{15.8} & 12.3 & 14.2 & \multicolumn{1}{c|}{11.8} & 13.1 & 14.3 & 12.8 \\
DeepSeek V2 Lite Chat & 12.6 & 13.4 & \multicolumn{1}{c|}{11.4} & 9.8 & 10.5 & \multicolumn{1}{c|}{9.0} & 9.4 & 10.2 & \multicolumn{1}{c|}{8.7} & 32.1 & 39.3 & \multicolumn{1}{c|}{29.0} & 29.5 & 36.5 & \multicolumn{1}{c|}{26.7} & 29.0 & 35.9 & 26.3 \\
DeepSeek Coder V2 Lite Instruct & 34.4 & 39.1 & \multicolumn{1}{c|}{32.4} & 34.6 & 39.2 & \multicolumn{1}{c|}{32.6} & 35.2 & 39.8 & \multicolumn{1}{c|}{33.1} & 64.8 & 72.8 & \multicolumn{1}{c|}{61.4} & 65.7 & 73.4 & \multicolumn{1}{c|}{62.5} & 65.4 & 73.4 & 62.0 \\
Gemma 3 4B IT & 35.8 & 37.9 & \multicolumn{1}{c|}{33.5} & 38.2 & 40.0 & \multicolumn{1}{c|}{36.3} & 37.2 & 39.6 & \multicolumn{1}{c|}{35.3} & 67.8 & 72.0 & \multicolumn{1}{c|}{65.1} & 68.2 & 72.5 & \multicolumn{1}{c|}{65.6} & 68.2 & 72.9 & 65.9 \\
Gemma 3 12B IT & \underline{ 48.4} & 52.8 & \multicolumn{1}{c|}{46.0} & 47.8 & 51.5 & \multicolumn{1}{c|}{45.2} & 47.4 & \underline{ 51.1} & \multicolumn{1}{c|}{44.8} & \underline{ 76.6} & \underline{ 82.1} & \multicolumn{1}{c|}{\underline{ 73.8}} & 75.0 & 80.8 & \multicolumn{1}{c|}{72.4} & \underline{ 75.5} & \underline{ 81.2} & 71.3 \\
Gemma 3 27B IT & \textbf{53.0} & \textbf{55.2} & \multicolumn{1}{c|}{\textbf{52.9}} & \textbf{51.6} & \textbf{54.5} & \multicolumn{1}{c|}{\textbf{50.5}} & \textbf{52.6} & \textbf{55.0} & \multicolumn{1}{c|}{\textbf{50.4}} & \textbf{77.6} & \textbf{82.7} & \multicolumn{1}{c|}{\textbf{74.6}} & \textbf{76.4} & \underline{ 81.4} & \multicolumn{1}{c|}{\textbf{73.4}} & \textbf{76.7} & \textbf{81.7} & \textbf{73.6} \\
Granite 3.2 8B Instruct & 29.0 & 33.6 & \multicolumn{1}{c|}{27.2} & 29.2 & 33.1 & \multicolumn{1}{c|}{27.2} & 29.0 & 33.3 & \multicolumn{1}{c|}{26.6} & 58.5 & 67.7 & \multicolumn{1}{c|}{55.1} & 59.7 & 67.7 & \multicolumn{1}{c|}{56.0} & 58.4 & 66.9 & 54.9 \\
Granite 3.3 8B Instruct & 28.6 & 32.1 & \multicolumn{1}{c|}{26.4} & 29.4 & 33.2 & \multicolumn{1}{c|}{27.6} & 30.4 & 33.8 & \multicolumn{1}{c|}{28.6} & 60.7 & 69.0 & \multicolumn{1}{c|}{57.2} & 60.5 & 69.3 & \multicolumn{1}{c|}{57.0} & 60.5 & 68.5 & 57.1 \\
Llama 3.1 8B Instruct & 33.2 & 36.4 & \multicolumn{1}{c|}{30.6} & 29.6 & 32.7 & \multicolumn{1}{c|}{27.0} & 28.6 & 32.3 & \multicolumn{1}{c|}{26.5} & 53.7 & 61.4 & \multicolumn{1}{c|}{50.0} & 52.4 & 60.1 & \multicolumn{1}{c|}{48.4} & 53.5 & 60.3 & 49.6 \\
Ministral 8B Instruct & 36.0 & 42.2 & \multicolumn{1}{c|}{33.7} & 36.8 & 42.9 & \multicolumn{1}{c|}{34.4} & 36.0 & 42.2 & \multicolumn{1}{c|}{33.7} & 57.6 & 67.0 & \multicolumn{1}{c|}{53.8} & 57.1 & 66.4 & \multicolumn{1}{c|}{53.5} & 55.5 & 64.4 & 52.1 \\
Mistral Small 24B Instruct & 39.2 & 47.0 & \multicolumn{1}{c|}{36.4} & 35.6 & 43.5 & \multicolumn{1}{c|}{33.1} & 35.2 & 42.8 & \multicolumn{1}{c|}{32.7} & 59.8 & 68.5 & \multicolumn{1}{c|}{54.6} & 54.0 & 62.8 & \multicolumn{1}{c|}{50.0} & 52.7 & 61.0 & 48.8 \\
Qwen 2.5 7B Instruct & 39.4 & 43.9 & \multicolumn{1}{c|}{36.7} & 39.0 & 43.8 & \multicolumn{1}{c|}{36.4} & 39.6 & 43.9 & \multicolumn{1}{c|}{36.6} & 66.6 & 74.1 & \multicolumn{1}{c|}{63.2} & 66.4 & 73.5 & \multicolumn{1}{c|}{62.8} & 65.9 & 73.2 & 62.6 \\
Qwen 2.5 14B Instruct & 40.6 & 45.9 & \multicolumn{1}{c|}{38.6} & 38.2 & 43.5 & \multicolumn{1}{c|}{35.9} & 38.4 & 44.1 & \multicolumn{1}{c|}{36.4} & 67.8 & 76.4 & \multicolumn{1}{c|}{64.2} & 67.4 & 75.7 & \multicolumn{1}{c|}{64.0} & 66.2 & 74.9 & 62.5 \\
Qwen 2.5 32B Instruct & 26.0 & 31.6 & \multicolumn{1}{c|}{24.7} & 27.6 & 32.9 & \multicolumn{1}{c|}{26.3} & 23.6 & 27.8 & \multicolumn{1}{c|}{22.6} & 69.5 & 78.2 & \multicolumn{1}{c|}{66.6} & 69.1 & 77.8 & \multicolumn{1}{c|}{65.9} & 68.3 & 77.0 & 65.3 \\
Qwen 2.5 Coder 7B Instruct & 44.8 & 49.3 & \multicolumn{1}{c|}{42.5} & 44.2 & 49.2 & \multicolumn{1}{c|}{41.9} & 43.0 & 48.6 & \multicolumn{1}{c|}{40.9} & 74.5 & 81.4 & \multicolumn{1}{c|}{71.8} & 74.0 & \textbf{81.5} & \multicolumn{1}{c|}{71.0} & 72.1 & 79.5 & 69.6 \\
Qwen 2.5 Coder 14B Instruct & 31.4 & 35.8 & \multicolumn{1}{c|}{30.3} & 41.0 & 45.9 & \multicolumn{1}{c|}{38.9} & 39.0 & 44.0 & \multicolumn{1}{c|}{36.9} & 74.8 & 80.8 & \multicolumn{1}{c|}{72.1} & \underline{ 75.5} & \underline{ 81.4} & \multicolumn{1}{c|}{\underline{ 72.8}} & 74.3 & 80.5 & \underline{ 71.4} \\
Qwen 2.5 Coder 32B Instruct & 48.0 & \underline{ 53.9} & \multicolumn{1}{c|}{\underline{ 47.6}} & \underline{ 48.2} & \underline{ 54.1} & \multicolumn{1}{c|}{\underline{ 46.3}} & \underline{ 48.4} & \textbf{55.0} & \multicolumn{1}{c|}{\underline{ 47.2}} & 68.8 & 76.5 & \multicolumn{1}{c|}{65.7} & 60.7 & 65.8 & \multicolumn{1}{c|}{58.4} & 58.2 & 62.9 & 56.0 \\
StarCoder 15B Instruct & 30.6 & 36.1 & 28.9 & 30.0 & 35.7 & 28.6 & 30.8 & 36.4 & 29.3 & 59.9 & 69.1 & 57.2 & 60.6 & 69.6 & 58.1 & 61.0 & 70.2 & 58.3 \\
\bottomrule
\end{tabular}
}
\end{center}
\end{table*}

%% file: tables/plan-res.tex
\begin{table*}[ht]
    \centering
    \caption{
    Performance of the Planner–Coder pipeline on the \textbf{BIRD} and \textbf{Spider} datasets. 
    \textbf{Bold} marks the best performance, and \underline{underlined} denotes the second-best. 
    Overall, \textbf{Planner–Coder conditioning consistently improves SQL generation accuracy across both datasets}, with the strongest gains observed for weaker or mid-tier coders guided by large or joint planner agents.}
    \label{tbl:planner-coder-bird-spider}
    \resizebox{0.95\textwidth}{!}{%
        \begin{tabular}{l ccc | ccc | ccc | ccc | ccc}
            \toprule
            
            & \multicolumn{15}{c}{\textbf{BIRD}} \\ \midrule
            \textbf{Model} & \multicolumn{3}{c | }{\textbf{R1-7B}} & \multicolumn{3}{c | }{\textbf{R1-14B}} & \multicolumn{3}{c | }{\textbf{R1-32B}} & \multicolumn{3}{c |}{\textbf{QwQ-32B}} & \multicolumn{3}{c}{\textbf{R1-32B + QwQ-32B}} \\
            & \textbf{EX} & \textbf{Soft F1} & \textbf{VES} & \textbf{EX} & \textbf{Soft F1} & \textbf{VES} & \textbf{EX} & \textbf{Soft F1} & \textbf{VES} & \textbf{EX} & \textbf{Soft F1} & \textbf{VES} & \textbf{EX} & \textbf{Soft F1} & \textbf{VES} \\
            \midrule
                CodeLlama Instruct 7B & 7.8 & 9.4 & \multicolumn{1}{c|}{7.3} & 9.4 & 12.0 & \multicolumn{1}{c|}{9.1} & 13.0 & 15.3 & \multicolumn{1}{c|}{12.6} & 17.8 & 20.7 & \multicolumn{1}{c|}{16.9} & 30.0 & 35.4 & 28.8 \\
                CodeLlama Instruct 13B & 7.8 & 8.8 & \multicolumn{1}{c|}{7.3} & 13.4 & 17.6 & \multicolumn{1}{c|}{12.9} & 18.6 & 22.3 & \multicolumn{1}{c|}{17.8} & 26.8 & 31.3 & \multicolumn{1}{c|}{25.5} & 30.8 & 35.1 & 29.5 \\
                CodeLlama Instruct 34B & 20.2 & 23.5 & \multicolumn{1}{c|}{18.9} & 9.4 & 13.4 & \multicolumn{1}{c|}{7.1} & 15.6 & 18.7 & \multicolumn{1}{c|}{13.3} & 22.4 & 26.3 & \multicolumn{1}{c|}{19.1} & 33.0 & 38.3 & 29.6 \\
                DeepSeek V2 Lite Chat & 14.4 & 17.7 & \multicolumn{1}{c|}{13.7} & 25.2 & 30.0 & \multicolumn{1}{c|}{23.6} & 28.6 & 33.8 & \multicolumn{1}{c|}{27.1} & 34.0 & 38.6 & \multicolumn{1}{c|}{32.2} & 35.2 & 40.6 & 33.4 \\
                DeepSeek Coder V2 Lite Instruct & 22.8 & 27.5 & \multicolumn{1}{c|}{21.6} & 37.8 & 44.1 & \multicolumn{1}{c|}{36.0} & 39.8 & 46.8 & \multicolumn{1}{c|}{38.1} & 45.0 & 51.6 & \multicolumn{1}{c|}{43.1} & 43.8 & 50.9 & 41.8 \\
                Gemma 3 4B IT & 32.0 & 34.5 & \multicolumn{1}{c|}{30.5} & 43.2 & 46.8 & \multicolumn{1}{c|}{41.0} & 46.2 & 51.0 & \multicolumn{1}{c|}{44.3} & 43.8 & 48.0 & \multicolumn{1}{c|}{41.6} & 48.2 & 52.8 & 46.3 \\
                Gemma 3 12B IT & 40.0 & 43.7 & \multicolumn{1}{c|}{38.2} & \underline{ 49.2} & \underline{ 53.4} & \multicolumn{1}{c|}{\underline{ 46.7}} & \underline{ 50.2} & \underline{ \textbf{55.6}} & \multicolumn{1}{c|}{\underline{ 48.0}} & \underline{ 50.6} & 56.5 & \multicolumn{1}{c|}{\underline{ 48.0}} & \underline{ 51.4} & \underline{ 56.5} & \underline{ 49.1} \\
                Gemma 3 27B IT & \textbf{46.4} & \textbf{49.0} & \multicolumn{1}{c|}{\textbf{44.0}} & \textbf{51.4} & \textbf{55.8} & \multicolumn{1}{c|}{\textbf{49.8}} & \textbf{51.4} & \underline{ 55.4} & \multicolumn{1}{c|}{\textbf{48.8}} & \textbf{56.0} & \textbf{60.4} & \multicolumn{1}{c|}{\textbf{53.4}} & \textbf{56.4} & \textbf{60.2} & \textbf{50.3} \\
                Granite 3.2 8B Instruct & 17.2 & 20.2 & \multicolumn{1}{c|}{16.3} & 36.8 & 42.2 & \multicolumn{1}{c|}{34.9} & 42.2 & 49.6 & \multicolumn{1}{c|}{40.5} & 42.8 & 49.7 & \multicolumn{1}{c|}{40.6} & 48.2 & 55.3 & 46.0 \\
                Granite 3.3 8B Instruct & 21.8 & 25.3 & \multicolumn{1}{c|}{20.8} & 40.0 & 45.8 & \multicolumn{1}{c|}{37.8} & 42.8 & 49.9 & \multicolumn{1}{c|}{41.3} & 48.6 & 55.4 & \multicolumn{1}{c|}{46.2} & 46.8 & 52.9 & 44.9 \\
                Llama 3.1 8B Instruct & 23.6 & 25.9 & \multicolumn{1}{c|}{22.2} & 32.0 & 36.6 & \multicolumn{1}{c|}{30.5} & 28.2 & 33.4 & \multicolumn{1}{c|}{27.3} & 40.4 & 47.0 & \multicolumn{1}{c|}{38.4} & 43.8 & 50.6 & 41.5 \\
                Ministral 8B Instruct & 22.8 & 26.6 & \multicolumn{1}{c|}{21.4} & 39.6 & 45.0 & \multicolumn{1}{c|}{37.6} & 40.8 & 48.4 & \multicolumn{1}{c|}{39.1} & 46.4 & 54.3 & \multicolumn{1}{c|}{44.3} & 45.4 & 52.7 & 43.4 \\
                Mistral Small 24B Instruct & 39.2 & \underline{ 45.9} & \multicolumn{1}{c|}{36.5} & 44.2 & 50.7 & \multicolumn{1}{c|}{42.0} & 40.0 & 47.1 & \multicolumn{1}{c|}{38.2} & 46.0 & 52.9 & \multicolumn{1}{c|}{43.3} & 45.6 & 51.9 & 43.5 \\
                Qwen 2.5 7B Instruct & 25.6 & 30.6 & \multicolumn{1}{c|}{24.0} & 41.6 & 47.9 & \multicolumn{1}{c|}{39.4} & 43.0 & 50.3 & \multicolumn{1}{c|}{40.7} & 45.2 & 52.2 & \multicolumn{1}{c|}{43.1} & 45.8 & 53.0 & 43.7 \\
                Qwen 2.5 14B Instruct & 37.0 & 44.0 & \multicolumn{1}{c|}{34.8} & 44.8 & 50.9 & \multicolumn{1}{c|}{42.2} & 45.2 & 52.6 & \multicolumn{1}{c|}{43.3} & 49.0 & \underline{ 56.7} & \multicolumn{1}{c|}{46.6} & 47.2 & 54.4 & 45.0 \\
                Qwen 2.5 32B Instruct & \underline{ 42.0} & 43.7 & \multicolumn{1}{c|}{\underline{ 40.3}} & 43.4 & 45.6 & \multicolumn{1}{c|}{41.2} & 47.2 & 49.1 & \multicolumn{1}{c|}{44.4} & 48.0 & 50.4 & \multicolumn{1}{c|}{46.1} & 48.2 & 50.1 & 45.3 \\
                Qwen 2.5 Coder 7B Instruct & 22.2 & 25.9 & \multicolumn{1}{c|}{21.2} & 37.2 & 42.7 & \multicolumn{1}{c|}{35.4} & 42.4 & 49.4 & \multicolumn{1}{c|}{40.5} & 41.4 & 48.2 & \multicolumn{1}{c|}{39.6} & 44.8 & 52.3 & 42.9 \\
                Qwen 2.5 Coder 14B Instruct & 28.0 & 33.1 & \multicolumn{1}{c|}{26.5} & 37.0 & 41.7 & \multicolumn{1}{c|}{35.2} & 35.2 & 41.4 & \multicolumn{1}{c|}{33.6} & 44.0 & 51.3 & \multicolumn{1}{c|}{41.7} & 44.8 & 52.2 & 42.7 \\
                Qwen 2.5 Coder 32B Instruct & 34.2 & 36.3 & \multicolumn{1}{c|}{30.2} & 36.4 & 38.4 & \multicolumn{1}{c|}{31.5} & 42.0 & 47.9 & \multicolumn{1}{c|}{39.0} & 35.2 & 39.1 & \multicolumn{1}{c|}{33.5} & 40.8 & 44.7 & 36.3 \\
                StarCoder 15B Instruct & 10.6 & 12.8 & \multicolumn{1}{c|}{10.0} & 18.2 & 20.5 & \multicolumn{1}{c|}{17.0} & 15.0 & 16.7 & \multicolumn{1}{c|}{14.3} & 23.4 & 26.8 & \multicolumn{1}{c|}{22.4} & 26.6 & 30.7 & 25.5 \\
            \midrule

            & \multicolumn{15}{c}{\textbf{SPIDER}} \\
            \midrule
            CodeLlama Instruct 13B & 34.2 & 39.4 & \multicolumn{1}{c|}{32.9} & 36.7 & 41.6 & \multicolumn{1}{c|}{35.4} & 45.2 & 50.2 & \multicolumn{1}{c|}{43.3} & 51.5 & 58.5 & \multicolumn{1}{c|}{49.4} & 56.2 & 63.0 & 54.0 \\
            CodeLlama Instruct 34B & 39.1 & 44.8 & \multicolumn{1}{c|}{37.2} & 38.6 & 45.0 & \multicolumn{1}{c|}{36.9} & 47.2 & 54.0 & \multicolumn{1}{c|}{45.5} & 50.4 & 56.5 & \multicolumn{1}{c|}{48.4} & 56.1 & 63.8 & 53.8 \\
            CodeLlama Instruct 7B & 31.0 & 35.3 & \multicolumn{1}{c|}{30.0} & 30.2 & 35.0 & \multicolumn{1}{c|}{29.0} & 38.0 & 43.4 & \multicolumn{1}{c|}{36.6} & 41.2 & 47.3 & \multicolumn{1}{c|}{39.6} & 51.2 & 59.1 & 49.1 \\
            DeepSeek Coder V2 Lite Instruct & 58.4 & 66.9 & \multicolumn{1}{c|}{55.2} & 63.6 & 72.6 & \multicolumn{1}{c|}{60.8} & 67.3 & 76.6 & \multicolumn{1}{c|}{64.5} & 66.9 & 75.4 & \multicolumn{1}{c|}{63.9} & 67.3 & 75.6 & 64.6 \\
            DeepSeek V2 Lite Chat & 43.6 & 50.1 & \multicolumn{1}{c|}{41.4} & 51.2 & 59.7 & \multicolumn{1}{c|}{48.8} & 52.1 & 59.5 & \multicolumn{1}{c|}{49.8} & 56.6 & 64.3 & \multicolumn{1}{c|}{54.1} & 62.2 & 70.0 & 59.3 \\
            Gemma 3 4B IT & 60.4 & 66.7 & \multicolumn{1}{c|}{57.8} & 69.1 & 75.2 & \multicolumn{1}{c|}{\underline{ 66.3}} & 69.6 & 76.8 & \multicolumn{1}{c|}{66.8} & 70.8 & 77.0 & \multicolumn{1}{c|}{68.3} & 72.1 & 78.7 & \underline{ 69.6} \\
            Gemma 3 12B IT & \underline{ 70.0} & \underline{ 75.2} & \multicolumn{1}{c|}{\underline{ 66.7}} & \underline{ 72.7} & \underline{ 79.0} & \multicolumn{1}{c|}{\textbf{70.1}} & \textbf{73.8} & \textbf{81.1} & \multicolumn{1}{c|}{\textbf{71.0}} & \textbf{73.5} & \textbf{80.2} & \multicolumn{1}{c|}{\textbf{71.1}} & \textbf{72.8} & \textbf{79.9} & \textbf{70.2} \\
            Gemma 3 27B IT & \textbf{71.2} & \textbf{77.3} & \multicolumn{1}{c|}{68.2} & \textbf{73.0} & \textbf{79.4} & \multicolumn{1}{c|}{\textbf{70.1}} & \underline{ 73.3} & \underline{ 80.5} & \multicolumn{1}{c|}{\underline{ 69.9}} & \underline{ 73.0} & \underline{ 78.9} & \multicolumn{1}{c|}{\underline{ 70.8}} & \underline{ 71.7} & \underline{ 79.5} & 68.9 \\
            Granite 3.2 8B Instruct & 51.5 & 58.6 & \multicolumn{1}{c|}{48.8} & 63.8 & 73.2 & \multicolumn{1}{c|}{60.8} & 67.3 & 76.9 & \multicolumn{1}{c|}{64.6} & 65.8 & 73.5 & \multicolumn{1}{c|}{63.6} & 68.0 & 77.0 & 65.2 \\
            Granite 3.3 8B Instruct & 52.9 & 60.0 & \multicolumn{1}{c|}{50.1} & 64.8 & 74.4 & \multicolumn{1}{c|}{61.7} & 67.2 & 77.0 & \multicolumn{1}{c|}{64.3} & 68.7 & 76.9 & \multicolumn{1}{c|}{66.0} & 67.9 & 76.9 & 65.1 \\
            Llama 3.1 8B Instruct & 49.8 & 57.4 & \multicolumn{1}{c|}{47.3} & 57.8 & 65.8 & \multicolumn{1}{c|}{55.2} & 57.7 & 66.5 & \multicolumn{1}{c|}{55.1} & 64.2 & 72.8 & \multicolumn{1}{c|}{61.4} & 65.6 & 73.9 & 63.1 \\
            Ministral 8B Instruct & 54.8 & 62.9 & \multicolumn{1}{c|}{51.9} & 64.3 & 73.1 & \multicolumn{1}{c|}{61.0} & 63.9 & 73.0 & \multicolumn{1}{c|}{61.0} & 67.7 & 75.8 & \multicolumn{1}{c|}{65.2} & 68.6 & 76.9 & 65.8 \\
            Mistral Small 24B Instruct & 61.2 & 71.2 & \multicolumn{1}{c|}{58.1} & 62.3 & 70.0 & \multicolumn{1}{c|}{59.1} & 60.2 & 68.7 & \multicolumn{1}{c|}{57.6} & 68.6 & 76.8 & \multicolumn{1}{c|}{66.1} & 67.5 & 76.0 & 64.6 \\
            Qwen 2.5 7B Instruct & 59.1 & 66.9 & \multicolumn{1}{c|}{56.0} & 66.2 & 74.9 & \multicolumn{1}{c|}{63.0} & 67.3 & 76.6 & \multicolumn{1}{c|}{64.5} & 65.4 & 73.5 & \multicolumn{1}{c|}{62.9} & 67.5 & 75.7 & 64.7 \\
            Qwen 2.5 14B Instruct & 63.5 & 71.8 & \multicolumn{1}{c|}{60.1} & 68.5 & 77.0 & \multicolumn{1}{c|}{64.9} & 68.3 & 77.1 & \multicolumn{1}{c|}{65.4} & 69.1 & 76.7 & \multicolumn{1}{c|}{66.7} & 68.3 & 76.2 & 65.4 \\
            Qwen 2.5 32B Instruct & 62.6 & 72.1 & \multicolumn{1}{c|}{59.6} & 64.4 & 73.4 & \multicolumn{1}{c|}{61.5} & 62.6 & 71.4 & \multicolumn{1}{c|}{60.0} & 64.9 & 72.6 & \multicolumn{1}{c|}{62.5} & 64.3 & 71.5 & 62.1 \\
            Qwen 2.5 Coder 7B Instruct & 51.7 & 59.0 & \multicolumn{1}{c|}{49.1} & 60.7 & 69.7 & \multicolumn{1}{c|}{57.8} & 63.2 & 71.9 & \multicolumn{1}{c|}{60.7} & 60.2 & 68.0 & \multicolumn{1}{c|}{57.7} & 64.5 & 72.9 & 62.0 \\
            Qwen 2.5 Coder 14B Instruct & 53.2 & 59.4 & \multicolumn{1}{c|}{50.4} & 56.9 & 65.2 & \multicolumn{1}{c|}{54.4} & 55.1 & 63.4 & \multicolumn{1}{c|}{52.9} & 62.5 & 69.4 & \multicolumn{1}{c|}{60.3} & 65.7 & 73.6 & 62.9 \\
            Qwen 2.5 Coder 32B Instruct & 58.4 & 67.1 & \multicolumn{1}{c|}{55.4} & 60.3 & 68.9 & \multicolumn{1}{c|}{57.7} & 58.9 & 67.4 & \multicolumn{1}{c|}{56.2} & 55.6 & 62.4 & \multicolumn{1}{c|}{53.5} & 59.6 & 66.7 & 57.4 \\
            StarCoder 15B Instruct & 37.0 & 41.8 & 35.5 & 40.5 & 45.6 & 38.7 & 41.8 & 46.1 & 40.4 & 43.2 & 48.4 & 41.6 & 47.7 & 52.5 & 46.0 \\
            \bottomrule
        \end{tabular}
    }
\end{table*}

%% file: tables/pick-res.tex
\begin{table*}[ht]
    \caption{Execution Accuracy (EX) Scores for the Coder-Aggregator pipeline. Size of coders refers to relative parameter counts of the models used to generate initial SQLs using zero-shot prompts (Figure~\ref{fig:zeroshot_prompt}) for the Aggregator (Figure~\ref{fig:picker_prompt}). 'Small' models: Qwen2.5-7B-Instruct, Qwen2.5-Coder-7B-Instruct, and Gemma 3 4B IT. 'Mid' models: Qwen2.5-14B-Instruct, Qwen2.5-Coder-14B-Instruct, and Gemma 3 12B IT. 'Large' models: Qwen2.5-32B-Instruct, Qwen2.5-Coder-32B-Instruct, and Gemma 3 27B IT. \textbf{Bold} indicates the best model, \underline{underlined} the second-best.}
    \label{tbl:pick-res}
    \resizebox{0.8\textwidth}{!}{
    \begin{tabular}{l|ccccccccc}
        \hline
         & \multicolumn{3}{c |}{\textbf{SMALL}} & \multicolumn{3}{c|}{\textbf{MID}} & \multicolumn{3}{c}{\textbf{LARGE}} \\
         \textbf{Model} & \textbf{EX} & \textbf{Soft F1} & \multicolumn{1}{c|}{\textbf{R-VES}} & \textbf{EX} & \textbf{Soft F1} & \multicolumn{1}{c|}{\textbf{R-VES}} & \textbf{EX} & \textbf{Soft F1} & \textbf{R-VES} \\ \hline
         & \multicolumn{9}{c}{\textbf{BIRD}} \\ \hline
        DeepSeek R1 Distill Qwen 7B Instruct & 36.4 & 40.9 & \multicolumn{1}{c|}{34.7} & 41.6 & 47.2 & \multicolumn{1}{c|}{39.4} & 45.8 & 52.1 & 43.6 \\
        DeepSeek R1 Distill Qwen 14B Instruct & \underline{ 47.8} & \underline{ 52.9} & \multicolumn{1}{c|}{\underline{ 45.9}} & 48.2 & 53.2 & \multicolumn{1}{c|}{45.9} & 49.6 & 57.0 & 47.4 \\
        DeepSeek R1 Distill Qwen 32B Instruct & 46.2 & 50.2 & \multicolumn{1}{c|}{44.8} & \underline{ 50.4} & \underline{ 56.4} & \multicolumn{1}{c|}{\underline{ 48.0}} & \underline{ 50.4} & \underline{ 58.04} & \underline{ 47.5} \\
        QwQ 32B & \textbf{51.0} & \textbf{54.7} & \multicolumn{1}{c|}{\textbf{48.9}} & \textbf{53.2} & \textbf{58.0} & \multicolumn{1}{c|}{\textbf{48.9}} & \textbf{54.4} & \textbf{60.2} & \textbf{53.1} \\ \midrule
         & \multicolumn{9}{c}{\textbf{SPIDER}} \\ \midrule
        DeepSeek R1 Distill Qwen 7B Instruct & 64.5 & 72.2 & \multicolumn{1}{c|}{61.7} & 67.8 & 75.4 & \multicolumn{1}{c|}{64.9} & 69.4 & 77.4 & 66.7 \\
        DeepSeek R1 Distill Qwen 14B Instruct & \textbf{73.8} & \textbf{81.2} & \multicolumn{1}{c|}{\textbf{71.2}} & \textbf{74.6} & \textbf{81.5} & \multicolumn{1}{c|}{\textbf{72.3}} & \textbf{75.1} & \textbf{82.2} & \textbf{72.4} \\
        DeepSeek R1 Distill Qwen 32B Instruct & \underline{ 70.9} & \underline{ 78.7} & \multicolumn{1}{c|}{\underline{ 68.7}} & \underline{ 73.1} & \underline{ 80.2} & \multicolumn{1}{c|}{\underline{ 70.6}} & \underline{ 72.1} & \underline{ 78.7} & \underline{ 69.0} \\
        QwQ 32B & 67.6 & 74.0 & \multicolumn{1}{c|}{65.5} & 65.5 & 70.9 & \multicolumn{1}{c|}{63.2} & 63.9 & 70.1 & 60.8 \\ \hline
    \end{tabular}
    }
\end{table*}

%% file: sections/appendix.tex
\clearpage
\begin{center}
{\Large \textbf{Appendix}}
\end{center}

\section{Prompts}
\label{sec:prompts}
This section presents the prompts used in our three proposed pipelines:
(1) the Multi-Agent Discussion pipeline, where agents iteratively critique and improve SQL queries, and a judge synthesizes the final answer;
(2) the Planner-Coder pipeline, where a planning model outlines the reasoning steps and a coding model generates the SQL query; and
(3) the Coder-Aggregator pipeline, where multiple coder agents generate queries independently, and a reasoning agent selects or synthesizes the final output.

\subsection{Zero-Shot Prompts}
We first establish a baseline by using a single-turn, zero-shot prompt, shown in Figure~\ref{fig:zeroshot_prompt}.
\input{prompts/zs-prompt}

\subsection{Multi-Agent Discussion Prompts}
The Multi-Agent Discussion pipeline involves three agent roles: Starter, Discussion, and Judge. 
Figure~\ref{fig:mad_starter_prompt} shows the prompt used by the Starter Agent to generate initial responses. 
Figure~\ref{fig:mad_discuss_prompt} shows the prompt for the Discussion Agent to critique and revise responses. 
Figure~\ref{fig:mad_judge_prompt} presents the prompt used by the Judge Agent to synthesize the final SQL query.
\input{prompts/mad-prompt}

\subsection{Planner-Coder Prompts}
The Planner-Coder pipeline first uses the prompt in Figure~\ref{fig:pc_planner_prompt} to generate a step-by-step plan for answering the question. 
This plan is then passed to the Coder Agent, which uses the prompt in Figure~\ref{fig:pc_coder_prompt} to produce the final SQL query.
\input{prompts/plan-prompt}

\subsection{Coder-Aggregator Prompts}
The Coder-Aggregator pipeline begins with multiple coders generating candidate SQL queries independently. 
Figure~\ref{fig:picker_prompt} shows the prompt used by the Aggregator Agent to select or synthesize the best final query from these candidates.
\input{prompts/pick-prompt}

\section{Additional Figures}

\begin{figure*}[h]
    \centering
    \includegraphics[width=0.9\textwidth]{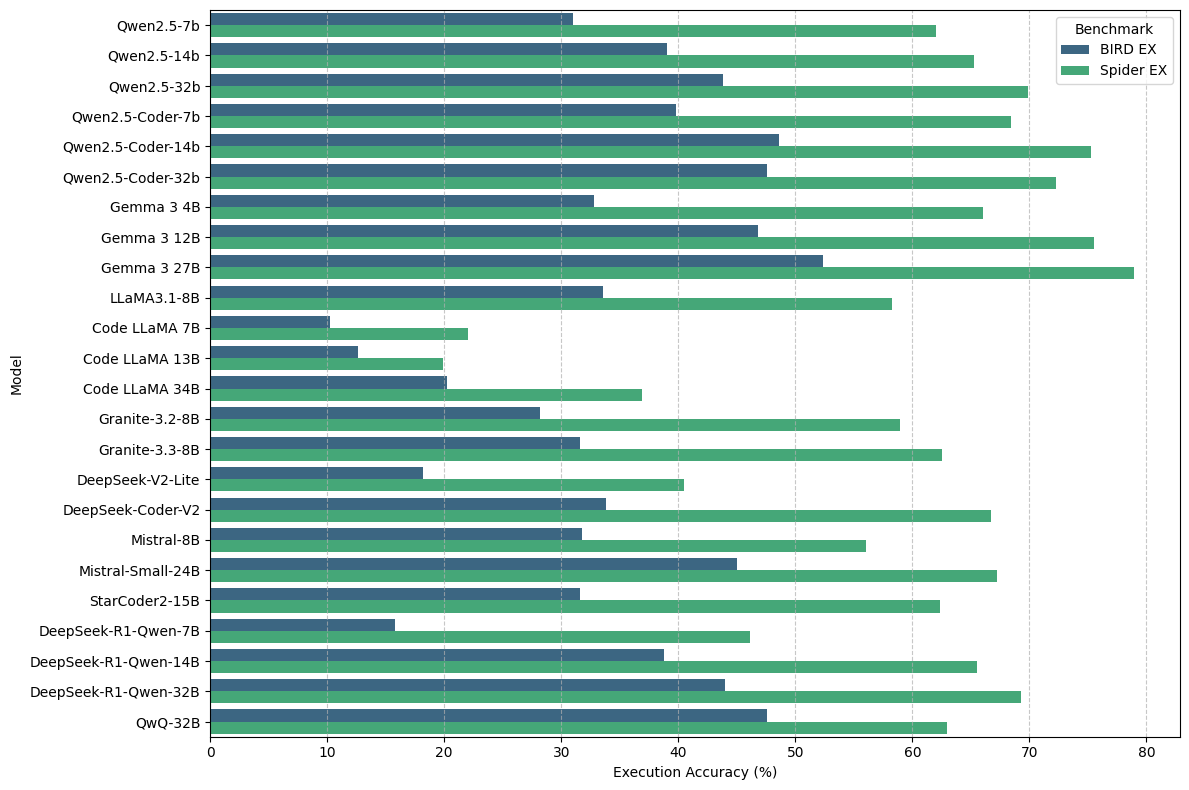}
    \caption{Comparison of the zeroshot EX score across all tested models on BIRD Mini-Dev and Spider Dev datasets.}
    \label{fig:zeroshot}
\end{figure*}

\begin{figure*}[h]
    \centering
    \includegraphics[width=0.9\textwidth]{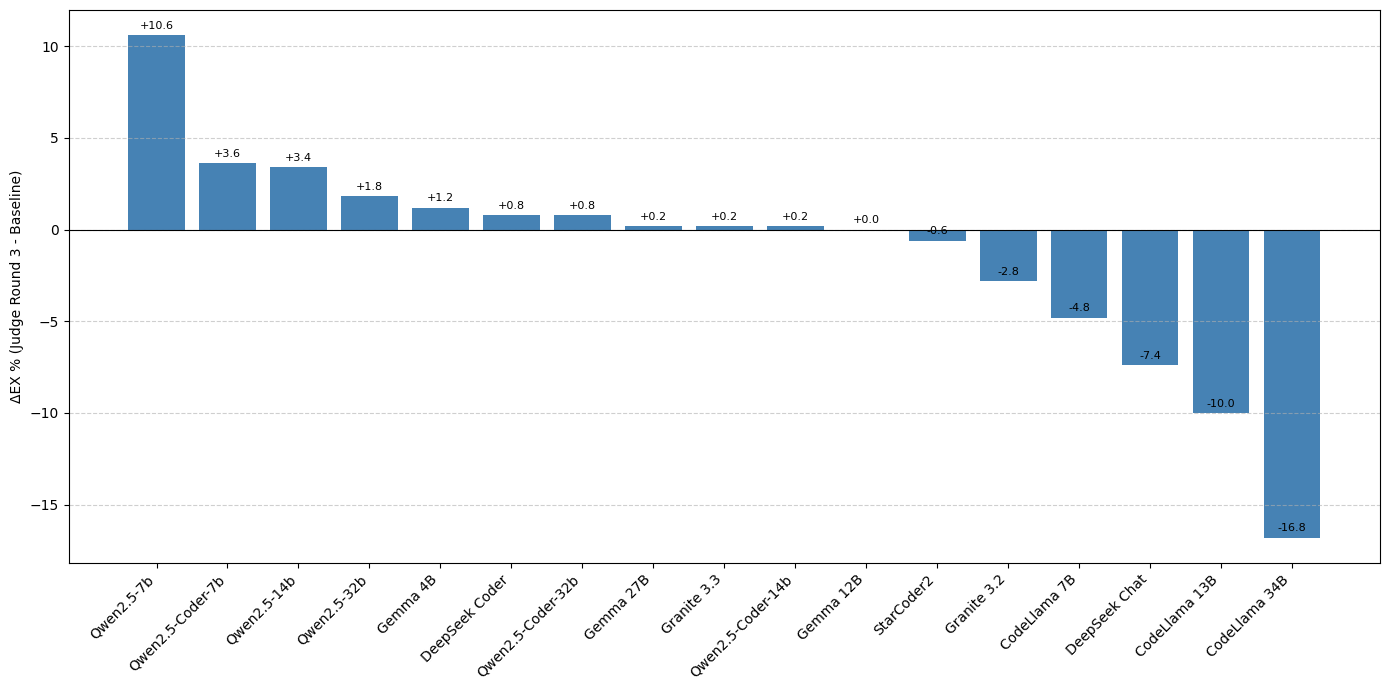}
    \caption{Change in judge EX score from baseline to the end of round three of multi-agent discussion across all tested models on BIRD Mini-Dev. The Qwen2.5 family of models derive the greatest benefit from multi-agent discussion.}
    \label{fig:delta}
\end{figure*}

In Figure \ref{fig:zeroshot} we visualize zeroshot performance of all 24 models we tested on the BIRD Mini-Dev and Spider Dev datasets. In Figure \ref{fig:delta} we see the relative performance delta of all 20 non-reasoning LLMs from their baseline to after 3 rounds of discussion. We calculate the difference as:

$\Delta = EX_{judge\_round3} - EX_{baseline}$

%% file: prompts/zs-prompt.tex
\begin{figure*}
    \begin{tcolorbox}[
        title=\textbf{Baseline Prompt},
        colback=lightgray,
        colframe=gray!60!black,
        boxrule=0.4pt,
        arc=2pt,
        width=0.8\linewidth,
        fonttitle=\bfseries\footnotesize,
        sharp corners,
        left=3pt,
        right=3pt,
        top=3pt,
        bottom=3pt,
        boxsep=1pt,
        nobeforeafter,
        center,
        toptitle=1pt,
        bottomtitle=1pt
    ]
    \begin{quote}
        \footnotesize{
            Given the following SQLite tables, your job is to write a query to answer the given question.
            
            \texttt{\#\#\# SCHEMA} \\
            \texttt{\{schema\}}
            
            \texttt{\#\#\# QUESTION} \\ 
            \texttt{\{question\}}
            
            \texttt{\#\#\# YOUR RESPONSE} \\
            \texttt{Let's think step by step}
        }
    \end{quote}
    \end{tcolorbox}
    \caption{Zero-shot baseline prompt. The "Let's think step by step" portion is removed for reasoning models according to best practices outlined by model publishers.}
    \label{fig:zeroshot_prompt}
    \vspace{15pt}
\end{figure*}

%% file: prompts/mad-prompt.tex
\begin{figure*}[ht]
    \begin{tcolorbox}[
        title=\textbf{Starter Agent Prompt},
        colback=lightgray,
        colframe=gray!60!black,
        boxrule=0.4pt,
        arc=2pt,
        width=0.8\linewidth,
        fonttitle=\bfseries\footnotesize,
        sharp corners,
        left=3pt,
        right=3pt,
        top=3pt,
        bottom=3pt,
        boxsep=1pt,
        nobeforeafter,
        center,
        toptitle=1pt,
        bottomtitle=1pt
    ]
    \begin{quote}
        \footnotesize{
            You are a helpful SQL coding assistant who \{persona\}. Given the following SQLite tables, your job is to write a query to answer the given question.
            
            \texttt{\#\#\# SCHEMA} \\
            \texttt{\{schema\}}
            
            \texttt{\#\#\# QUESTION} \\ 
            \texttt{\{question\}}
            
            \texttt{\#\#\# YOUR RESPONSE} \\
            \texttt{Let's think step by step}
        }
    \end{quote}
    \end{tcolorbox}
    \caption{Zero-shot prompt used by the Starter Agent in the Multi-Agent Discussion pipeline. Each agent, assigned a unique persona, generates an initial SQL query. These starter responses are then  reviewed by neighboring agents in the first discussion round.}
    \label{fig:mad_starter_prompt}
    \vspace{15pt}
\end{figure*}

\begin{figure*}[ht]
    \begin{tcolorbox}[
        title=\textbf{Discussion Agent Prompt},
        colback=lightgray,
        colframe=gray!60!black,
        boxrule=0.4pt,
        arc=2pt,
        width=0.8\linewidth,
        fonttitle=\bfseries\footnotesize,
        sharp corners,
        left=3pt,
        right=3pt,
        top=3pt,
        bottom=3pt,
        boxsep=1pt,
        nobeforeafter,
        center,
        toptitle=1pt,
        bottomtitle=1pt
    ]
    \begin{quote}
        \footnotesize{
            You are a helpful coding agent. You are collaborating with several other coding agents to answer a given user question.

            Given the following SQLite tables, your job is to write a query to answer the given question. Please use the other agents' responses as additional information. Feel free to offer helpful suggestions and corrections to their answers as you see fit.

            \texttt{\#\#\# SCHEMA} \\
            \texttt{\{schema\}}
            
            \texttt{\#\#\# QUESTION} \\
            \texttt{\{question\}}

            \texttt{\#\#\# AGENT RESPONSES} \\
            \texttt{\#\#\#\#\#\# Agent 1} \\
            \texttt{\{agent\_1\_response\}} \\
            \texttt{\#\#\#\#\#\# Agent 3} \\
            \texttt{\{agent\_3\_response\}}
            
            \texttt{\#\#\# YOUR RESPONSE} \\
            \texttt{Let's think step by step}
        }
    \end{quote}
    \end{tcolorbox}
    \caption{Prompt given to Discussion Agents in the Multi-Agent Discussion pipeline. Each agent considers the responses of others across multiple rounds to refine its own SQL query. This example shows the prompt for Discussion Agent 2, incorporating responses from Agents 1 and 3. Final responses from all agents are passed to the Judge Agent (Figure~\ref{fig:mad_judge_prompt}) to generate the final SQL query.}
    \label{fig:mad_discuss_prompt}
    \vspace{15pt}
\end{figure*}

\begin{figure*}[ht]
    \begin{tcolorbox}[
        title=\textbf{Judge Agent Prompt},
        colback=lightgray,
        colframe=gray!60!black,
        boxrule=0.4pt,
        arc=2pt,
        width=0.8\linewidth,
        fonttitle=\bfseries\footnotesize,
        sharp corners,
        left=3pt,
        right=3pt,
        top=3pt,
        bottom=3pt,
        boxsep=1pt,
        nobeforeafter,
        center,
        toptitle=1pt,
        bottomtitle=1pt
    ]
    \begin{quote}
        \footnotesize{
            You are a SQL expert overseeing 3 coding agents collaborating to answer a user question.

            Given the following SQLite tables, their job is to write queries given a user’s request. As the expert, your job is to judge the merit of their work and combine their responses to generate a final production-ready SQLite query.

            \texttt{\#\#\# SCHEMA} \\
            \texttt{\{schema\}}
            
            \texttt{\#\#\# QUESTION} \\
            \texttt{\{question\}}

            \texttt{\#\#\# AGENT RESPONSES} \\
            \texttt{\#\#\#\#\#\# Agent 1} \\
            \texttt{\{agent\_1\_response\}} \\
            \texttt{\#\#\#\#\#\# Agent 2} \\
            \texttt{\{agent\_2\_response\}} \\
            \texttt{\#\#\#\#\#\# Agent 3} \\
            \texttt{\{agent\_3\_response\}}
            
            \texttt{\#\#\# YOUR VERDICT} \\
            \texttt{Let's think step by step}
        }
    \end{quote}
    \end{tcolorbox}
    \caption{Prompt given to the Judge Agent in the Multi-Agent Discussion pipeline. Judge Agent reviews the outputs of all Discussion Agents (Figure~\ref{fig:mad_discuss_prompt}) after each round and produces the final SQL query.}
    \label{fig:mad_judge_prompt}
    \vspace{15pt}
\end{figure*}

%% file: prompts/plan-prompt.tex
\begin{figure*}[ht]
    \begin{tcolorbox}[
        title=\textbf{Planner Agent Prompt},
        colback=lightgray,
        colframe=gray!60!black,
        boxrule=0.4pt,
        arc=2pt,
        width=0.8\linewidth,
        fonttitle=\bfseries\footnotesize,
        sharp corners,
        left=3pt,
        right=3pt,
        top=3pt,
        bottom=3pt,
        boxsep=1pt,
        nobeforeafter,
        center,
        toptitle=1pt,
        bottomtitle=1pt
    ]
    \begin{quote}
        \footnotesize{
            You are an expert database engineer. Your job is to analyze the given schema, and construct a step-by-step plan on how to answer the given question. The tables, columns and operations you outline will be used by students to generate final SQLite query.

            \texttt{\#\#\# SCHEMA} \\
            \texttt{\{schema\}}
            
            \texttt{\#\#\# QUESTION} \\
            \texttt{\{question\}}
            
            \texttt{\#\#\# YOUR RESPONSE}
        }
    \end{quote}
    \end{tcolorbox}
    \caption{Prompt given to the Planner Agent in the Planner-Coder pipeline. The generated plan is later used by Coder Agents (Figure~\ref{fig:pc_coder_prompt}) to produce the final SQL query.}
    \label{fig:pc_planner_prompt}
    \vspace{15pt}
\end{figure*}

\begin{figure*}[ht]
    \begin{tcolorbox}[
        title=\textbf{Coder Agent Prompt},
        colback=lightgray,
        colframe=gray!60!black,
        boxrule=0.4pt,
        arc=2pt,
        width=0.8\linewidth,
        fonttitle=\bfseries\footnotesize,
        sharp corners,
        left=3pt,
        right=3pt,
        top=3pt,
        bottom=3pt,
        boxsep=1pt,
        nobeforeafter,
        center,
        toptitle=1pt,
        bottomtitle=1pt
    ]
    \begin{quote}
        \footnotesize{
            Based on the given schema and plan, generate a single SQLite query to answer the question.

            \texttt{\#\#\# SCHEMA} \\
            \texttt{\{schema\}}
            
            \texttt{\#\#\# QUESTION} \\
            \texttt{\{question\}}

            \texttt{\#\#\# PLAN} \\
            \texttt{\{plan\}}
            
            \texttt{\#\#\# YOUR RESPONSE} \\
            \texttt{Let's think step by step}
        }
    \end{quote}
    \end{tcolorbox}
    \caption{Prompt given to a Coder Agent in the Planner-Coder pipeline. Coder Agent generates the final SQL query conditioned on the plan from the Planner Agent (Figure~\ref{fig:pc_planner_prompt}). The “Let’s think step by step” instruction is omitted when the Coder is a reasoning model, which uses a \texttt{<think>} token to initiate internal reasoning.}
    \label{fig:pc_coder_prompt}
    \vspace{15pt}
\end{figure*}

%% file: prompts/pick-prompt.tex
\begin{figure*}[ht]
    \centering
    \begin{tcolorbox}[
        title=\textbf{Aggregator Agent Prompt},
        colback=lightgray,
        colframe=gray!60!black,
        boxrule=0.4pt,
        arc=2pt,
        width=0.8\linewidth,
        fonttitle=\bfseries\footnotesize,
        sharp corners,
        left=3pt,
        right=3pt,
        top=3pt,
        bottom=3pt,
        boxsep=1pt,
        nobeforeafter,
        center,
        toptitle=1pt,
        bottomtitle=1pt
    ]
    \begin{quote}
        \footnotesize{
            You are an expert database engineer. Based on the following SQLite tables, your job is to generate a single SQLite query to answer the given question. Junior members of your team have written you several candidate queries along with some of their reasonings. Consider their work when writing your answer.
            \vspace{6pt}

            \texttt{\#\#\# SCHEMA} \\
            \texttt{\{schema\}}

            \texttt{\#\#\# QUESTION} \\
            \texttt{\{question\}} \quad \texttt{Context:} \texttt{\{hint\}}

            \texttt{\#\#\# CANDIDATE QUERIES} \\
            \texttt{\{coder\_outputs\}}

            \texttt{\#\#\# YOUR RESPONSE}
        }
    \end{quote}
    \end{tcolorbox}
    \caption{Prompt given to the reasoning model acting as the Aggregator Agent in the Coder-Aggregator pipeline. \texttt{coder\_outputs} contains SQL candidates from multiple Coding Agents generated via zero-shot prompting (figure~\ref{fig:zeroshot_prompt}). The Aggregator Agent generates the final SQL query by considering the outputs from the coders.}
    \label{fig:picker_prompt}
    \vspace{15pt}
\end{figure*}